\documentclass[12pt]{article}

\usepackage{graphicx}
\graphicspath{{figures/}} % set a path
\usepackage{caption}
\usepackage{subcaption}
\usepackage{physics}
\usepackage{array,longtable} % tables
\usepackage{placeins}
\usepackage{booktabs}
\usepackage{amssymb}
\usepackage{amsmath}

\usepackage{jheppub}

\usepackage{orcidlink} % ORCID Link

\title{Prospective Prediction of OOD Degradation from Source-Side Training Dynamics}
\author{Sasha (Alexander) Monin~\orcidlink{0000-0001-9219-6474}}
\emailAdd{amonin@mailbox.sc.edu}
\affiliation{University of South Carolina \\
712 Main St, 404 \\
Columbia SC, 29208

}

\abstract{
We study whether persistent out-of-distribution (OOD) degradation can be predicted before it is directly observed using only source-side training dynamics. In a controlled shortcut-learning setting, a simple logistic regression predictor develops a clear prospective signal, while training time alone does not. Temporal summaries of the source-side quantities are substantially more informative than their current values. When transferred without additional training from a CNN to an MLP, confidence and entropy dynamics retain substantial predictive information. These results provide a proof of principle that source-side training dynamics can contain an early warning signal for future OOD failure.

} 

\hypersetup{
    pdftitle={Prospective Prediction of OOD Degradation from Source-Side Training Dynamics},
    pdfauthor={Sasha (Alexander) Monin},
    pdfkeywords={out-of-distribution degradation, shortcut learning, training dynamics}
}

\begin{document}

\maketitle

\section{Introduction}

The rapid progress of AI systems makes questions of safety and reliability increasingly important. For broad discussions of technical AI safety, see~\cite{Amodei2016Concrete,Hendrycks2021Unsolved}. One important source of reliability failures is distribution shift: models that perform well on the training distribution can suffer substantial degradation when evaluated out of distribution (OOD) under realistic shifts~\cite{koh2021wilds}. More generally, successful optimization on the training objective does not uniquely determine how a model will behave outside the training distribution. Many different solutions may achieve essentially the same performance on the training and validation data while relying on different predictive rules. This phenomenon was discussed in~\cite{damour2022underspecification} under the name of underspecification.

A closely related and extensively studied manifestation of this problem is shortcut learning, where a model relies on predictive features that work well on the training distribution but fail under distribution shift~\cite{geirhos2020shortcut}. Such behavior has been demonstrated in a variety of settings, including medical imaging, image segmentation, and vision-language representation learning~\cite{Brown2023ShorT,Lin2024Shortcut,Bleeker2024Reduce}.

Training dynamics can reveal how such dependencies are acquired. For example,~\cite{swayamdipta2020dataset} characterizes individual examples using the model's confidence in the correct class and its variability over training. More directly in the context of shortcut learning,~\cite{murali2023beyond} shows that harmful spurious features can be detected by observing the learning dynamics of the early layers of deep networks. The work in~\cite{qiu2024complexity} further shows that simpler or more strongly correlated spurious features can slow the subsequent learning of core features, and that the learning of spurious and core features does not always separate into distinct stages.

Other approaches address OOD failure in different ways. The OOD accuracy of a model can be estimated without OOD labels by using unlabeled samples from the OOD distribution~\cite{Garg2022Leveraging,Baek2022Agreement}. Baek et al. also estimate OOD performance across checkpoints along a training trajectory. These methods estimate the model's OOD performance at the time of evaluation and require access to samples from the OOD distribution. Other methods use the behavior of individual training examples to identify examples that should receive greater emphasis during debiasing~\cite{Nam2020Learning,Liu2021JustTrainTwice}. Their goal is to improve robustness by modifying the training procedure.

These approaches address either the detection and mitigation of spurious behavior or the estimation of current OOD performance. This leaves a different prospective question. Detecting that a model has started to use a spurious feature does not determine when its effect will appear as a persistent OOD performance gap. Similarly, estimating the current OOD performance does not predict whether such a degradation will emerge later. We therefore ask whether source-side training dynamics contain information about an approaching OOD degradation before that degradation is directly observed.

In this paper, we study this question in a controlled shortcut-learning setting. We train models on a source distribution containing a perfectly correlated shortcut and evaluate them on a clean OOD distribution in which the shortcut is absent. The clean OOD distribution is used to define the degradation event and in the trajectory collection and selection procedure, but is never provided to the predictor. At each training step, the predictor uses only source-side quantities observed up to that time. For many trajectories, the source and clean OOD accuracies evolve similarly for a substantial initial period, while a large and persistent gap only develops later. We ask whether the source-side information available before this event can be used to predict that the degradation is approaching.

We find that a simple logistic regression predictor can detect a prospective signal before the degradation event. This signal cannot be explained by training time alone and depends strongly on the recent history of the source-side quantities. For the CNN trajectories, temporal summaries of the training dynamics consistently perform better than the current values alone. Parameter and gradient dynamics alone also retain a prospective signal, showing that the effect is not confined to output-level performance statistics.

Finally, we test whether the same signal survives a change of model architecture. We train the predictor on the CNN trajectories and apply it without any additional training to an MLP with different degradation dynamics. The predictor using the complete set of source-side quantities transfers only weakly. However, when we restrict the predictor to confidence and entropy dynamics, a substantial predictive signal remains. This provides a proof of principle that \emph{source-side training dynamics can contain information about an upcoming OOD failure before this failure is directly observed}.

The code used in this work is publicly available~\cite{sasha2026code}.

\section{Experimental Setup}

\subsection{Shortcut-learning model organism}

We use a binary version of MNIST as a controlled shortcut-learning environment, assigning label $0$ to digits $0$-$4$ and label $1$ to digits $5$-$9$. The images are converted to RGB and a single label-correlated colored pixel is added in the upper-left corner. The resulting source distribution has perfect shortcut-label correlation. The OOD distribution consists of the corresponding clean images with the shortcut removed.

\begin{figure}[h!]
\centering
\includegraphics[width=13cm]{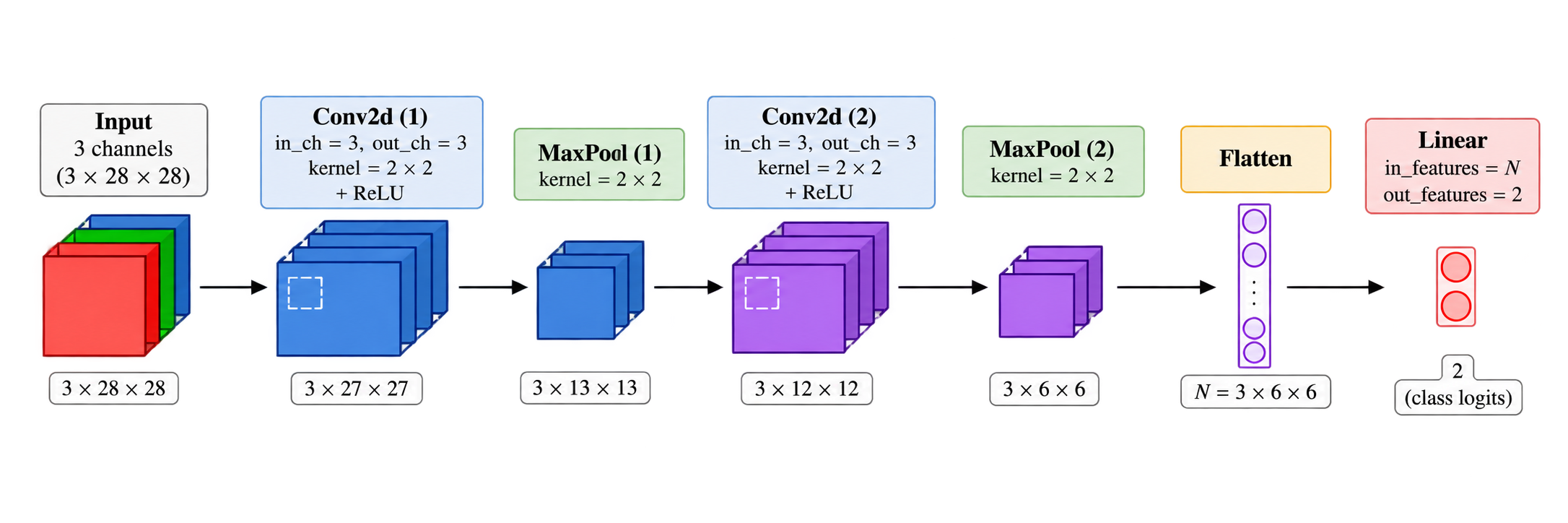}
\caption{Architecture of the CNN used as the main model organism.}
\label{fig:cnn_architecture}
\end{figure}

Our main model architecture is a small convolutional neural network (CNN), see Figure~\ref{fig:cnn_architecture}. It consists of two convolutional blocks, \texttt{Conv2d} + \texttt{ReLU} + \texttt{MaxPool2d}, followed by two fully connected layers. Both convolutional layers have three output channels, and all convolutional and max-pooling kernels have size $2\times2$. The hidden fully connected layer has dimension $4$, and the output layer has dimension $2$.

The main training parameters are summarized in Table~\ref{tab:training_setup}. Before introducing the shortcut, we verify that the architecture reaches at least $95\%$ accuracy on the clean dataset within $T_{\max}=12\,000$ training steps, corresponding to 10 epochs, see Appendix~\ref{app:clean_performance}. We then train multiple models with different random seeds on the shortcut dataset for at most $T_{\max}$ steps. A run is stopped earlier if, for 100 consecutive evaluation steps, $A_{\rm source}>0.99$ and $A_{\rm source}-A_{\rm clean}>0.45$.

\begin{table}[h!]
\centering
\small
\begin{tabular}{lc}
\hline
Training parameter & Value \\
\hline
Batch size $B$ & $50$ \\
Optimizer & Adam \\
Learning rate & $10^{-3}$ \\
$(\beta_1,\beta_2)$ & $(0.9,0.999)$ \\
Loss & Cross entropy \\
Maximum training steps $T_{\max}$ & $12\,000$ \\
\hline
\end{tabular}
\caption{Training parameters used throughout the experiments.}
\label{tab:training_setup}
\end{table}

\FloatBarrier

For each training trajectory, corresponding to a different random seed, we record the nine source-side quantities listed in Table~\ref{tab:source_observables}.
\begin{table}[h!]
\centering
\small
\begin{tabular}{ll}
\hline
Observable & Definition \\
\hline
Training loss & Cross-entropy loss on the current batch \\
Training accuracy & Accuracy on the current batch \\
Gradient norm & $\|\vec{g}_t\|$ \\
Parameter norm & $\|\vec{\theta}_t\|$ \\
Local parameter change & $\|\vec{\theta}_t-\vec{\theta}_{t-1}\|$ \\
Global parameter change & $\|\vec{\theta}_t-\vec{\theta}_0\|$ \\
Confidence & Mean maximum class probability \\
Confidence correct & Mean maximum class probability on correct predictions \\
Entropy & Mean prediction entropy \\
\hline
\end{tabular}
\caption{Source-side quantities recorded during training. Here $\vec{\theta}_t$ and $\vec{g}_t$ denote the concatenated parameter and gradient vectors at training step $t$.}
\label{tab:source_observables}
\end{table}

\subsection{Persistent degradation event}

We evaluate the source and clean accuracies after every training step and define the degradation by
\begin{equation}
D(t)=|A_{\rm source}(t)-A_{\rm clean}(t)|.
\end{equation}
The degradation time $T_{\rm deg}$ is the first time for which $D(T_{\rm deg})>\delta_{\rm deg}=0.05$ and at least 81 of the following 90 steps also satisfy this condition. This persistence requirement suppresses transient fluctuations. Information from the clean dataset after the candidate degradation time is used only offline to determine $T_{\rm deg}$ and is never provided to the predictor.

\subsection{Prospective prediction task}

Our main goal is to predict the degradation event using source-side training dynamics only. At every training step $t$, the predictor uses the preceding $W=100$ training steps, including the current step, and predicts whether persistent degradation will occur within a horizon $H=200$. A prediction at time $t$ is therefore considered positive when
\begin{equation}
0 < T_{\rm deg}-t \leq H.
\end{equation}

\subsection{Trajectory selection}

\begin{figure}[h!]
\centering
\includegraphics[width=8cm]{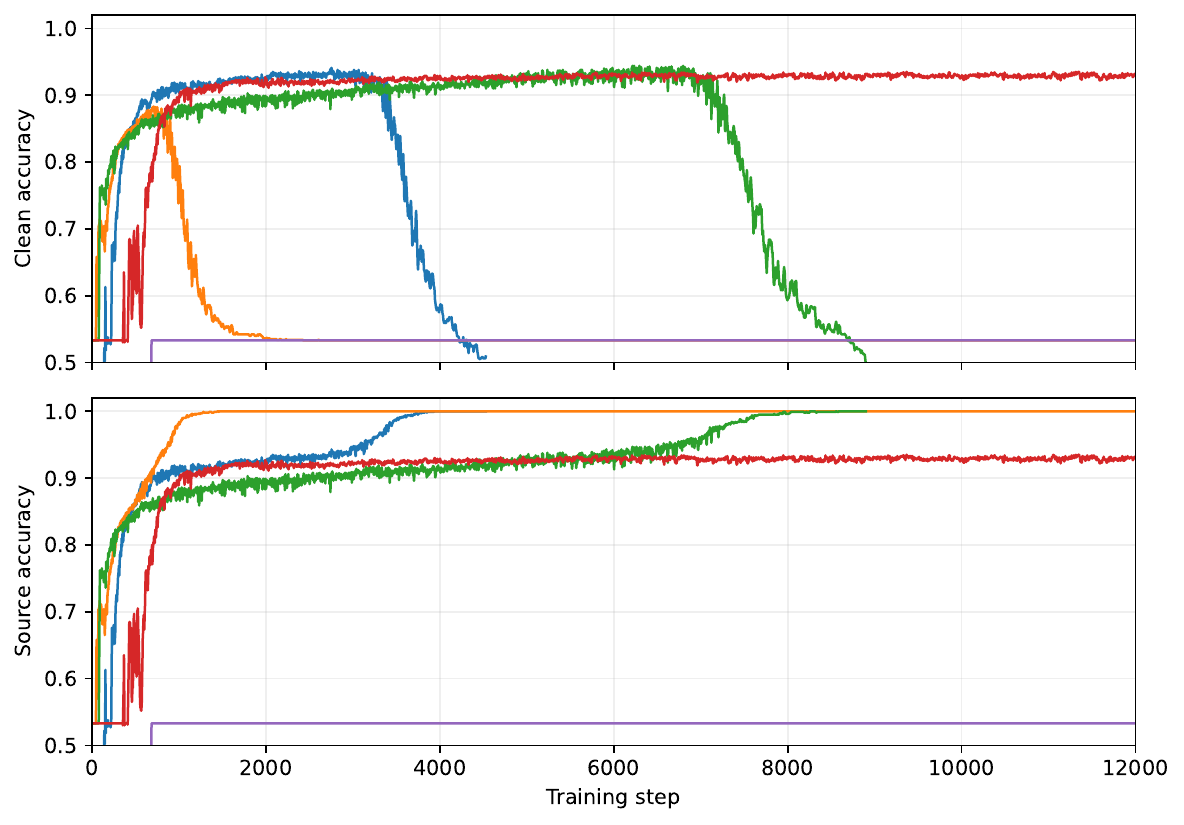}
\caption{Sample CNN training trajectories for different random seeds. Shown are the source and clean accuracies.}
\label{fig:shortcut_trajectories_cnn}
\end{figure}

Beyond the degradation-event definition and the stopping criterion described above, the OOD information is used here only to select trajectories for predictor training. Different random seeds can lead to substantially different OOD behavior, as illustrated in Figure~\ref{fig:shortcut_trajectories_cnn}, see also~\cite{damour2022underspecification}. We retain trajectories that reach $A_{\rm source}>0.75$, satisfy the persistent degradation criterion defining $T_{\rm deg}$, and have at least $S_{\rm pre}=W+2H=500$ pre-degradation steps. This allows us to construct a balanced set of positive windows within $H$ steps of $T_{\rm deg}$ and equally many earlier negative windows.

Trajectories that degrade too early are excluded as fast-degradation runs. Runs whose source accuracy does not exceed $A_{\rm stalled}=0.54$ are excluded as stalled. Trajectories with no observed degradation are not used to train the predictor, but are retained for the full-history evaluation of false warnings. For these trajectories, we include only windows satisfying $t+H\leq T_{\rm end}$.

The degrading trajectories are split into training and test sets before individual windows are extracted. Thus, no trajectory contributes windows to both sets. Consecutive windows share $W-1$ of their $W$ measurements and are therefore strongly correlated, so the trajectory is the relevant independent unit when comparing predictor variants.

\subsection{Source-side observables and predictor}

For each source-side quantity, we summarize the preceding $W=100$ steps by its current value, mean, standard deviation, and slope. With nine source-side quantities, each window is represented by $36$ features. Each feature is normalized using the mean and standard deviation computed on the predictor training set, and the same normalization is used for all test datasets.

We use a logistic regression predictor. If $\vec{x}(t)$ denotes the normalized feature vector for the window ending at time $t$, the predicted probability is
\begin{equation}
p(t)=\sigma\left(\vec{w}\cdot\vec{x}(t)+b\right).
\end{equation}
The predictor is trained on the positive and negative windows defined above using binary cross-entropy with logits and Adam with learning rate $10^{-3}$, $\beta_1=0.9$, $\beta_2=0.999$, for $7500$ optimization steps. For the classification metrics below and for the fraction of flagged trajectories shown in the figures, we classify a window as positive when $p(t)>0.5$.

\subsection{Architecture-transfer setting}

To test whether the prospective signal survives a change of architecture, we also consider a small MLP with two fully connected hidden layers of dimensions $4$ and $2$. As for the CNN, we verify that the architecture can reach approximately $95\%$ accuracy on the clean dataset, see Appendix~\ref{app:clean_performance}.

\begin{figure}[h!]
\centering
\includegraphics[width=8cm]{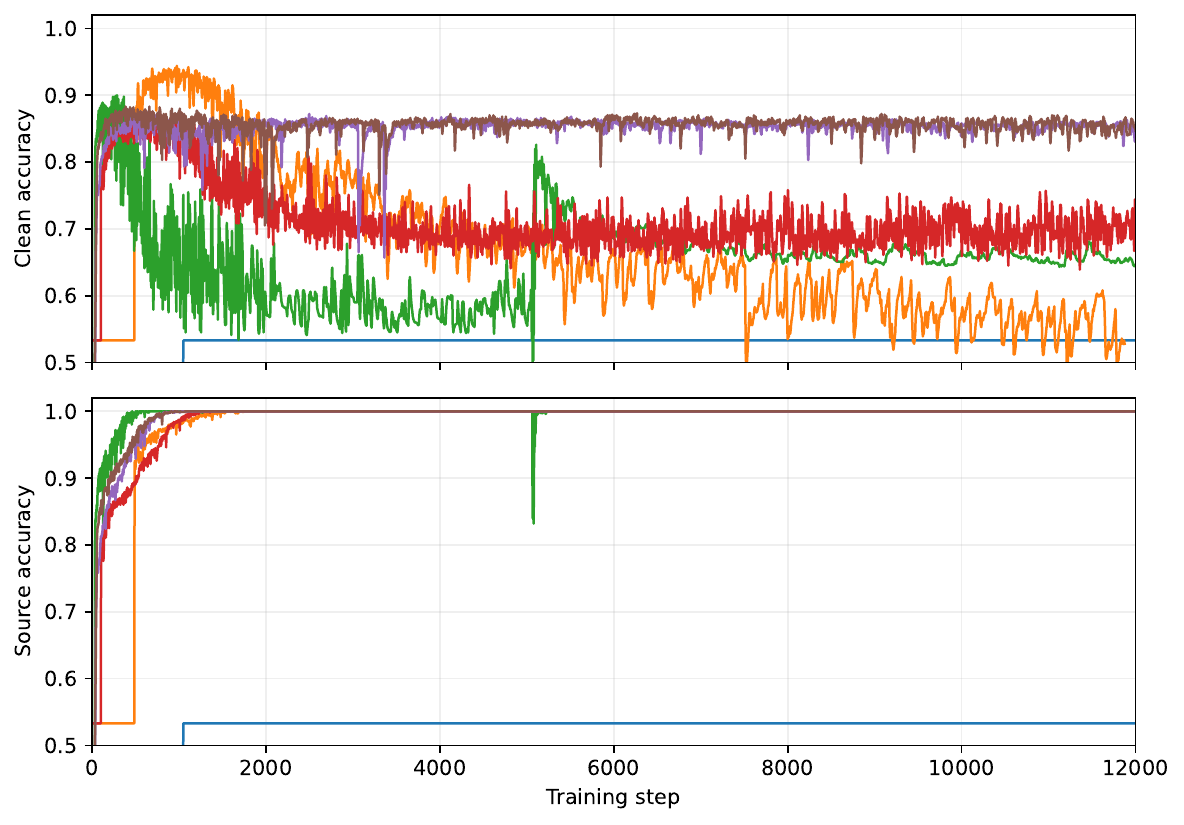}
\caption{Sample training trajectories for the MLP trained on the shortcut dataset. Different random seeds lead to different training behavior. Shown are the clean accuracy and the source accuracy.}
\label{fig:shortcut_trajectories_mlp}
\end{figure}

The MLP trajectories show qualitatively different degradation dynamics from the CNN trajectories, as illustrated in Figure~\ref{fig:shortcut_trajectories_mlp}. We use the same operational definition of the degradation event. For the MLP, this criterion generally identifies a persistent separation between the source and clean accuracies rather than a late collapse of the clean accuracy. The predictor trained on the CNN trajectories is then applied directly to the MLP trajectories without retraining.

\FloatBarrier

\section{Results}

In this section we present the main results. We first study prospective prediction within the CNN architecture and then test whether the learned signal generalizes to the MLP architecture without any additional retraining.

\subsection{Time-only baseline}

Before considering the source-side training dynamics, we ask whether the degradation event can be predicted from the training time alone. We train the same logistic predictor on the same balanced prediction dataset, using only the training step $t$ as input. This controls for the possibility that degradation simply tends to occur at a characteristic stage of training.

\begin{figure}[h!]
\begin{center}
\includegraphics[width=8cm]{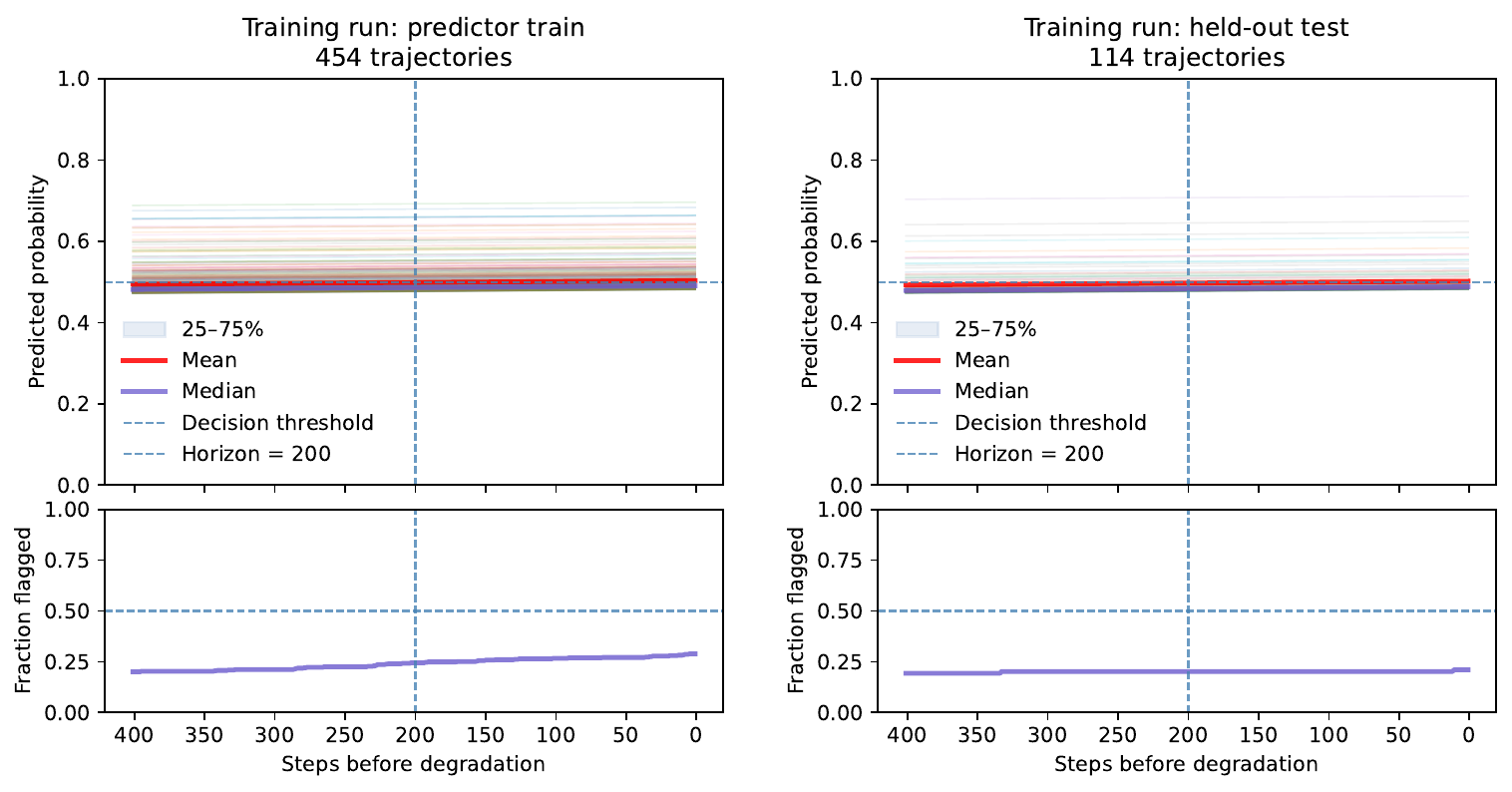}
\end{center}
\caption{Performance of the time-only predictor on the balanced prediction dataset. Shown are the predictor-training trajectories and the held-out test trajectories. The upper panels show the predicted probability as a function of the number of training steps before degradation, and the lower panels show the corresponding fraction of flagged trajectories.}
\label{fig:predictor_time_only}
\end{figure}

The result is shown in Figure~\ref{fig:predictor_time_only}. The time-only predictor does not distinguish windows far from and close to the degradation event. This is already the case on the trajectories used to train the predictor, and the behavior remains essentially unchanged on the held-out test trajectories. Thus, training time alone does not provide a useful prospective signal.

\FloatBarrier
\subsection{Prospective predictor training}

\begin{figure}[h!]
\begin{center}
\includegraphics[width=10cm]{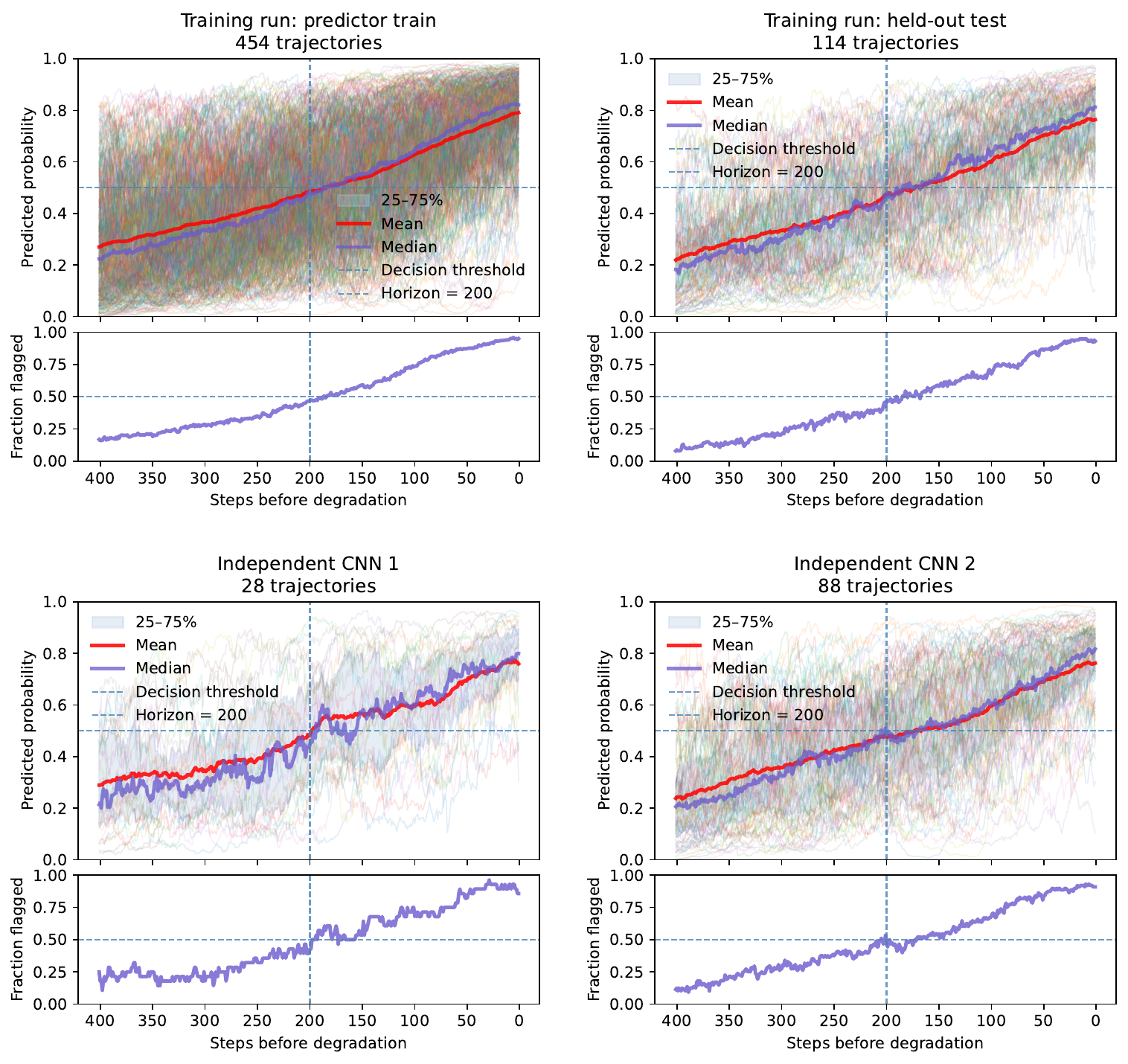}
\end{center}
\caption{Performance of the logistic regression predictor on the balanced prediction datasets. Shown are the predictor-training and held-out test trajectories from the training run, together with two independently generated CNN runs. The upper panels show the predicted probability as a function of the number of training steps before degradation, and the lower panels show the corresponding fraction of flagged trajectories.}
\label{fig:predictor_truncated}
\end{figure}

We now turn to the source-side training dynamics. We train the logistic regression predictor on the balanced prediction dataset constructed as described above. Figure~\ref{fig:predictor_truncated} shows its performance on the predictor-training and held-out test trajectories from the same CNN run, together with two independently generated CNN runs.

\begin{figure}[h!]
\begin{center}
\includegraphics[width=12cm]{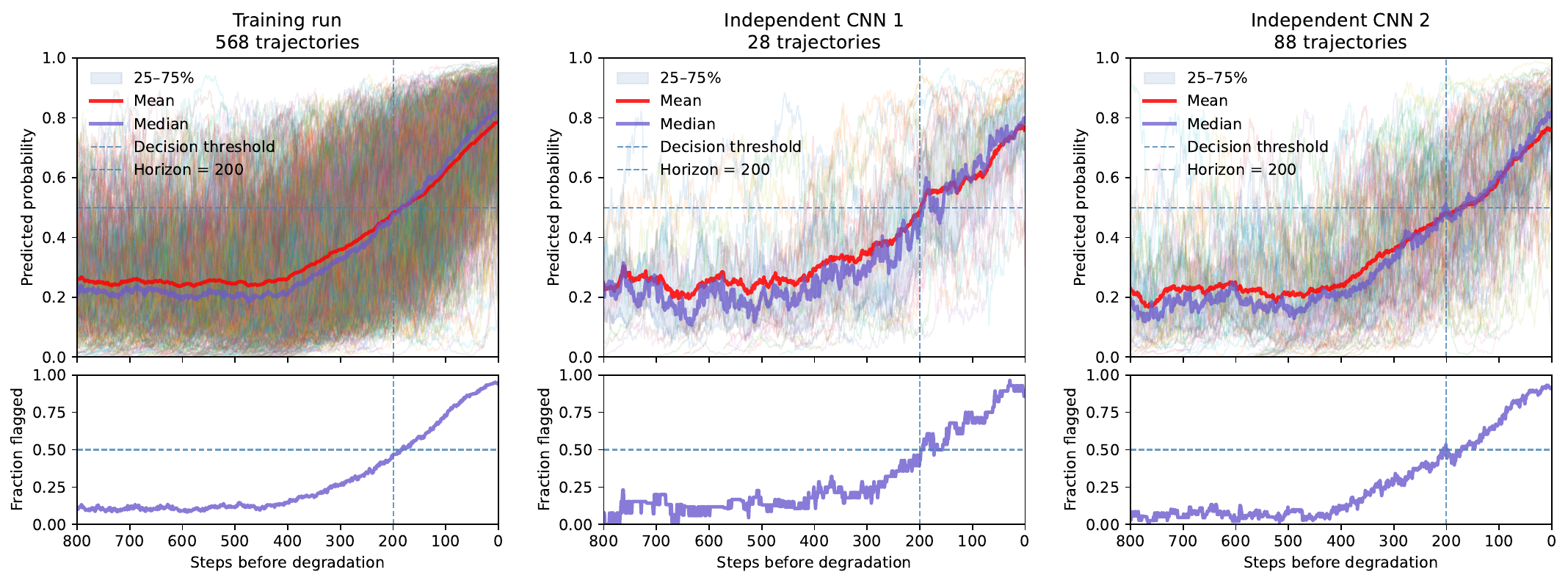}
\end{center}
\caption{Performance of the same trained predictor over the last $800$ training steps before degradation. Shown are all qualifying degrading trajectories from the training run, including those used to train the predictor, together with the same two independently generated CNN runs as in Figure~\ref{fig:predictor_truncated}. The upper panels show the predicted probability as a function of the number of training steps before degradation, while the lower panels show the corresponding fraction of flagged trajectories.}
\label{fig:predictor_full_history}
\end{figure}

We also examine the predictor over a wider part of the pre-degradation trajectory. At larger lead times fewer trajectories contribute to the aggregate statistics, making the curves progressively noisier. We therefore restrict this comparison to the last $800$ training steps before degradation. At this lead time approximately $40\%$-$45\%$ of the degrading trajectories still contribute in each of the three CNN runs. Figure~\ref{fig:predictor_full_history} shows the predictor over this wider range. For the training run, this extended diagnostic includes all qualifying degrading trajectories, including those used to train the predictor. The two independent CNN runs provide out-of-sample comparisons.

\FloatBarrier
\subsection{Metrics and features}
\label{sec:metrics_features}

We next investigate which source-side metrics and which features of their recent history are important for prospective prediction. Before comparing different choices, there is one immediate concern to address. Neither the source evaluation accuracy nor the clean evaluation accuracy is provided to the predictor, with the latter used only to define the degradation event. The predictor does, however, have access to the training loss and training accuracy. Since averaging the training accuracy over the prediction window strongly suppresses its batch-to-batch fluctuations, the predictor could in principle exploit a smoothed performance signal. For $B=50$ and $W=100$, the corresponding binomial standard deviation is at most approximately $0.007$.

We therefore perform a direct test by removing both the training loss and training accuracy from the predictor inputs. The results are summarized in Table~\ref{tab:metric_feature_ablations}. The predictive performance decreases only moderately and remains well above the time-only baseline. Thus, the prospective signal is not solely driven by the training loss and training accuracy as explicit inputs. This does not mean that all performance-related information has been removed. In binary classification, the remaining output statistics are themselves derived from the predicted class probabilities and can therefore retain information correlated with loss and accuracy, although this information is compressed by averaging over the batch.

\begin{table}[h]
\begin{center}
\small
\begin{tabular}{lccc}
\hline
Predictor inputs &
Training-run test &
Independent CNN 1 &
Independent CNN 2 \\
\hline
All, four features
    & 0.74 & 0.72 & 0.71 \\
All, mean
    & 0.71 & 0.70 & 0.70 \\    
All, current
    & 0.63 & 0.61 & 0.63 \\
\hline
\textbf{No loss/accuracy}
    & 0.71 & 0.69 & 0.67 \\
Output statistics
    & 0.66 & 0.67 & 0.64 \\    
Parameter/gradient
    & 0.65 & 0.63 & 0.67 \\
\hline
\end{tabular}
\end{center}
\caption{Prediction accuracy on the balanced prediction datasets for different choices of source-side metrics and temporal features. The datasets contain $114$, $28$, and $88$ trajectories for the Training-run test set, Independent CNN 1, and Independent CNN 2, respectively. ``All'' denotes all source-side metrics. ``Four features'' denotes the current value, mean, standard deviation, and slope over the preceding window. Unless stated otherwise, the four temporal features are used.}
\label{tab:metric_feature_ablations}
\end{table}

We next consider the role of the temporal features. Keeping all source-side metrics, the current value alone performs substantially worse than the mean over the preceding window. Adding the current value, standard deviation, and slope gives a further but smaller improvement. Thus, most of the improvement in balanced prediction accuracy comes from incorporating the recent history through the window mean.

The same distinction is particularly pronounced on trajectories with no observed degradation. Since all eligible windows on these trajectories are negative, they provide a direct test of false warnings. Table~\ref{tab:feature_false_positives} shows the mean trajectory-level false-positive rate for the same three choices of temporal features. For each trajectory, the false-positive rate is the fraction of eligible windows classified as positive, and we then average this quantity across trajectories.

\begin{table}[h!]
\begin{center}
\small
\begin{tabular}{lccc}
\hline
Features &
Training run &
Independent CNN 1 &
Independent CNN 2 \\
\hline
Four features
    & 0.03
    & 0.03
    & 0.03 \\
Mean
    & 0.09
    & 0.12
    & 0.13 \\
Current
    & 0.75
    & 0.82
    & 0.77 \\
\hline
\end{tabular}
\end{center}
\caption{Mean trajectory-level false-positive rates on trajectories with no observed degradation for different temporal features. The training run, Independent CNN 1, and Independent CNN 2 contain $84$, $6$, and $10$ such trajectories, respectively. For each trajectory, the false-positive rate is the fraction of eligible windows classified as positive, and the reported value is the mean across trajectories.}
\label{tab:feature_false_positives}
\end{table}

Using only the current values leads to false-positive rates between $0.75$ and $0.82$. Using the window mean reduces them to approximately $0.09$-$0.13$, while using all four temporal features reduces them further to approximately $0.03$. Thus, across all three CNN cohorts, temporal averaging greatly reduces false positives relative to the current-value predictor, and the additional temporal features reduce them further relative to the mean alone.

Finally, we separate the remaining metrics into two restricted groups. The output statistics, consisting of confidence, confidence on correctly classified examples, and entropy, and the parameter and gradient quantities both retain substantial predictive information, as shown in Table~\ref{tab:metric_feature_ablations}. Since the parameter and gradient group contains no output-level quantities, this shows that, within the CNN experiments, the prospective signal is not confined to output-level performance statistics. Their relative ordering changes between datasets, so we do not interpret one group as systematically more informative than the other. Neither group reproduces the performance obtained when all source-side metrics are combined. The corresponding prediction curves for the different choices of temporal features and source-side metrics on the balanced prediction datasets are shown in Appendix~\ref{app:predictor_tests}.

\subsection{Transfer to a different architecture}

We next test whether the prospective signal identified in the CNN trajectories survives a change of architecture. As discussed above, the MLP exhibits qualitatively different degradation dynamics. We use the same operational definition of the degradation event, which for the MLP generally identifies a persistent separation between the source and clean accuracies rather than a late collapse of the clean accuracy. In all cases, the predictor and its normalization are trained only on the CNN trajectories and are then applied directly to the MLP trajectories without additional training.

\begin{figure}[h]
\begin{center}
\begin{minipage}{0.32\textwidth}
\begin{center}
\includegraphics[width=\textwidth]{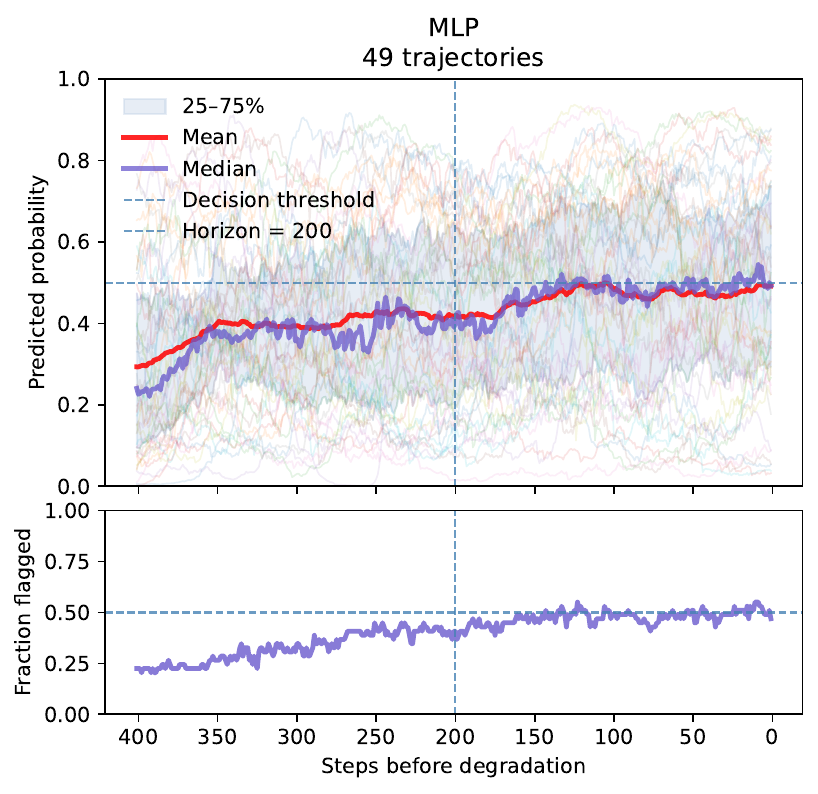}
\end{center}
\end{minipage}
\begin{minipage}{0.32\textwidth}
\begin{center}
\includegraphics[width=\textwidth]{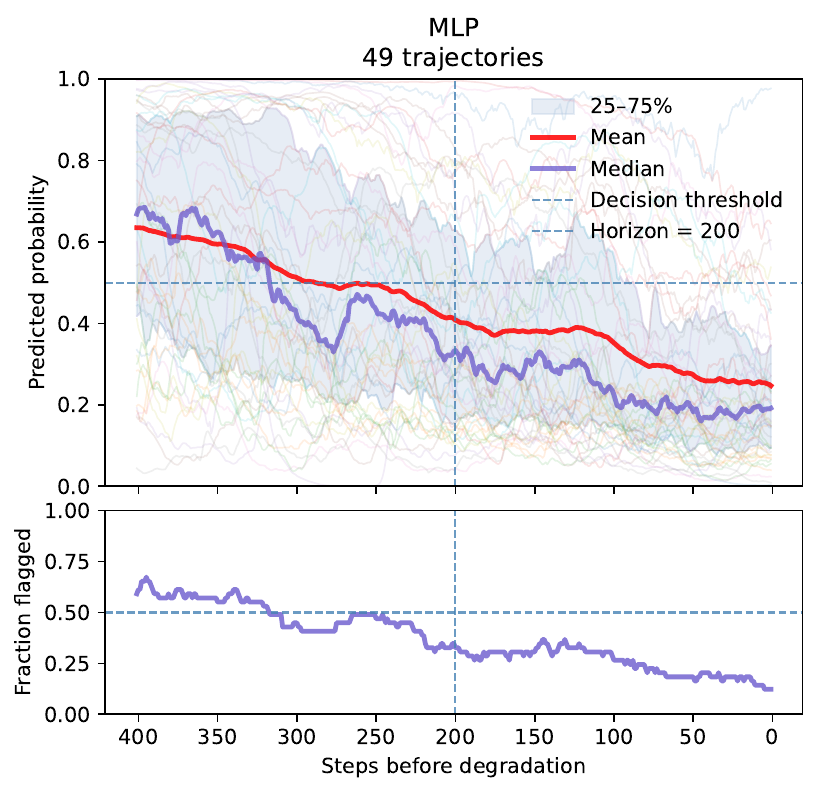}
\end{center}
\end{minipage}
\begin{minipage}{0.32\textwidth}
\begin{center}
\includegraphics[width=\textwidth]{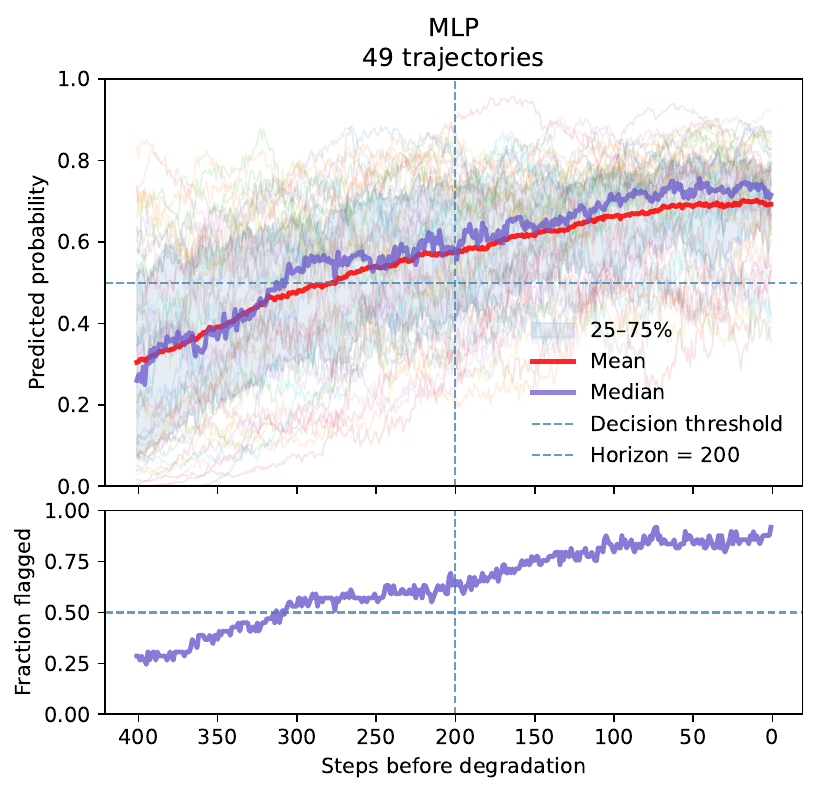}
\end{center}
\end{minipage}
\end{center}
\caption{Transfer of CNN-trained predictors to MLP trajectories without additional training. Left: predictor using all source-side quantities and the four temporal features. Center: predictor using only the parameter and gradient quantities. Right: predictor using only the output statistics. The complete predictor transfers only weakly, the parameter and gradient predictor transfers poorly, and the output statistics retain a substantially clearer prospective signal.}
\label{fig:mlp_transfer}
\end{figure}

We first use the complete set of source-side quantities and the same four temporal features as in the main CNN analysis. The prediction accuracy decreases from $0.74$ on the held-out CNN trajectories to $0.58$ on the balanced MLP dataset. As shown in the left panel of Figure~\ref{fig:mlp_transfer}, the predicted probability still tends to increase as the degradation event is approached, but the separation is substantially weaker than for the CNN trajectories. Thus, the complete predictor transfers only weakly to the MLP.

We then consider the two restricted metric groups introduced in Section~\ref{sec:metrics_features}. The parameter and gradient quantities depend directly on the architecture and parameterization of the model, so their numerical scales and training dynamics can change substantially between the CNN and MLP. Consistent with this, the predictor restricted to these quantities performs poorly on the MLP, with the prediction accuracy decreasing from $0.65$ on the held-out CNN trajectories to $0.38$ on the MLP.

The output statistics, consisting of confidence, confidence on correctly classified examples, and entropy, have the same interpretation for both architectures and do not depend directly on the parameterization of the network. The predictor restricted to these quantities transfers substantially better. Its prediction accuracy is $0.66$ on the held-out CNN trajectories and $0.65$ on the MLP. As shown in the right panel of Figure~\ref{fig:mlp_transfer}, the predicted probability also increases systematically as the degradation event is approached. The corresponding prediction accuracies, together with the mean-only and time-only comparisons, are summarized in Table~\ref{tab:mlp_transfer}.

\begin{table}[h]
\begin{center}
\small
\begin{tabular}{lcc}
\hline
Predictor & CNN held-out & MLP \\
\hline
All metrics, four features         & 0.74 & 0.58 \\
All metrics, mean                  & 0.71 & 0.61 \\
Output statistics, four features   & 0.66 & 0.65 \\
Parameter/gradient, four features  & 0.65 & 0.38 \\
Time only                          & 0.50 & 0.50 \\
\hline
\end{tabular}
\end{center}
\caption{Transfer of CNN-trained predictors to the MLP architecture. The predictors and their normalization are trained only on CNN trajectories and are applied to the MLP trajectories without additional training. The balanced evaluation datasets contain $114$ held-out CNN trajectories and $49$ MLP trajectories. Reported values are prediction accuracies on these datasets.}
\label{tab:mlp_transfer}
\end{table}

These results show that the source-side quantities are not equally transferable across architectures. The architecture-dependent parameter and gradient quantities transfer poorly, while the output statistics retain substantial predictive information across the change of architecture. Table~\ref{tab:mlp_transfer} also shows that, for the predictor using all source-side quantities, using only the window mean gives a somewhat higher MLP prediction accuracy, $0.61$ compared with $0.58$ for the four-feature predictor.

\section{Discussion}

The main result of this work is that persistent OOD degradation can be preceded by a detectable change in the source-side training dynamics. In our setting, a predictor using only information available up to the current training step can distinguish windows that are approaching the degradation event from windows that are still far from it. A time-only predictor does not reproduce this behavior, indicating that the signal is contained in the state and recent dynamics of the model rather than simply in the progress of training.

The temporal structure of the source-side quantities is important. Predictors using only their current values perform substantially worse than predictors that summarize their recent history. The window mean already captures most of the improvement in balanced prediction accuracy, while adding the current value, standard deviation, and slope gives only a modest further improvement. The clearer advantage of using all four temporal features appears on trajectories with no observed degradation, where they consistently produce fewer false positives than the mean alone.

Within the CNN experiments, the prospective signal is not confined to output-level statistics. The parameter and gradient quantities alone retain substantial predictive power. Removing the training loss and training accuracy also leaves a substantial signal, although the remaining output statistics can still contain loss-related information. The transfer experiment shows that the source-side quantities are not equally transferable across architectures. The full CNN predictor transfers poorly to the MLP, while the output statistics retain substantial predictive information across architectures.

There are several important limitations. The present experiments use small neural networks and a deliberately simple shortcut in binary MNIST that is perfectly correlated with the label. Whether similar prospective signals persist when this correlation is imperfect remains to be determined. In addition, the degradation time is defined operationally through the threshold $\delta_{\rm deg}=0.05$, so the timing of the prediction signal is relative to this definition. Choosing a smaller threshold would generally move $T_{\rm deg}$ earlier. The degradation event is also defined using performance on a clean OOD distribution. This information is used to define the event and select trajectories, but is never provided to the predictor. We therefore do not claim that the same source-side quantities will provide useful warning signals for arbitrary distribution shifts or larger models. Rather, the experiments establish in a controlled setting that \emph{source-side training dynamics can} contain information about a future degradation event before that event becomes visible in OOD performance.

A further limitation is that the MLP cohort contains no trajectories without observed degradation. The transfer experiment therefore tests whether the prospective signal survives a change of architecture. It does not allow us to establish a calibrated decision threshold or to estimate the false-positive rate on MLP trajectories with no observed degradation.

An important next step is to determine whether similar prospective signals can be identified in larger models and in settings where the shortcut and the relevant distribution shift are not specified in advance. Turning these prospective probabilities into an operational alarm is a separate problem that requires additional choices about acceptable warning times and false warnings, which we leave to future work.

\appendix

\section{Clean-dataset performance}
\label{app:clean_performance}

\begin{figure}[h]
\begin{center}

\begin{minipage}{0.48\textwidth}
\begin{center}
\includegraphics[width=\textwidth]{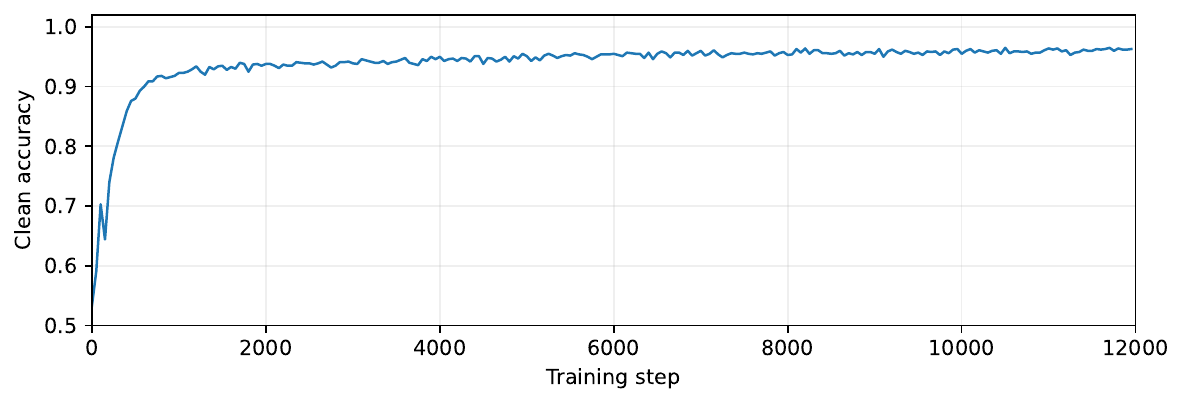}
\end{center}
\end{minipage}
\hfill
\begin{minipage}{0.48\textwidth}
\begin{center}
\includegraphics[width=\textwidth]{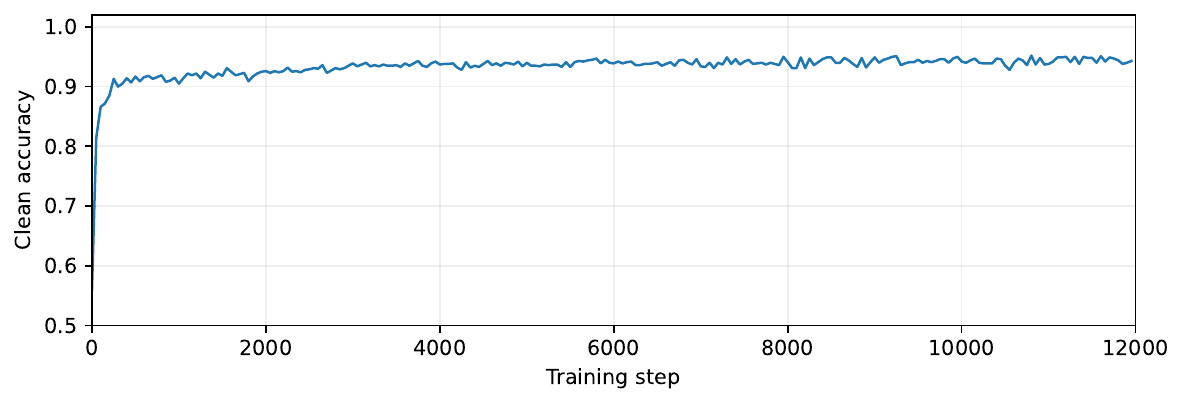}
\end{center}
\end{minipage}

\end{center}
\caption{Representative training trajectories on the clean dataset for the CNN (left) and MLP (right). The CNN reaches at least $95\%$ accuracy within the maximal training time $T_{\max}=12\,000$ steps, while the MLP reaches approximately $95\%$.}
\label{fig:clean_training}
\end{figure}

As a basic capacity check, before studying shortcut learning we verify that both model architectures are capable of learning the underlying classification task in the absence of the shortcut. Figure~\ref{fig:clean_training} shows representative clean-training trajectories for the CNN and MLP. These experiments are only used to establish that both architectures have sufficient capacity to solve the clean task.

\section{Additional predictor tests}
\label{app:predictor_tests}

In this section we present the prediction curves corresponding to the different choices of temporal features and source-side metrics discussed in Section~\ref{sec:metrics_features}. The numerical comparison is given in the main text. Here we show the corresponding predictor behavior on the balanced prediction datasets.

\begin{figure}[h!]
\begin{center}
\includegraphics[width=10cm]{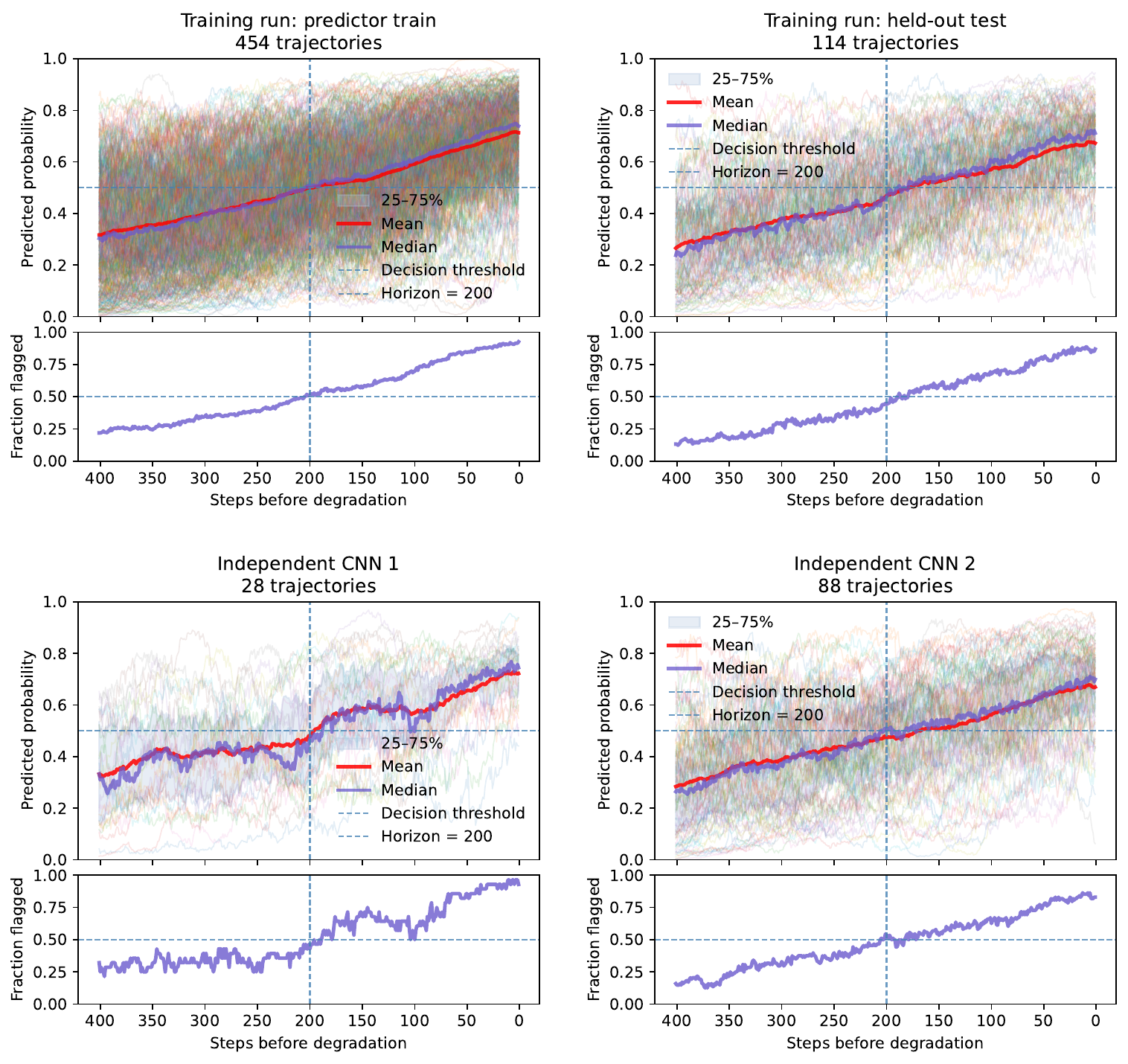}
\end{center}
\caption{Performance on the balanced prediction datasets after removing the training loss and training accuracy from the predictor inputs while retaining the four temporal features for the remaining source-side metrics.}
\label{fig:no_training_predictor_truncated}
\end{figure}

We first test the effect of removing the explicit training loss and training accuracy from the predictor inputs. Figure~\ref{fig:no_training_predictor_truncated} shows the result after removing both the training loss and training accuracy from the predictor inputs while retaining the four temporal features for all remaining source-side metrics.

\begin{figure}[h!]
\centering
\begin{minipage}{0.49\textwidth}
\centering
\includegraphics[width=\linewidth]{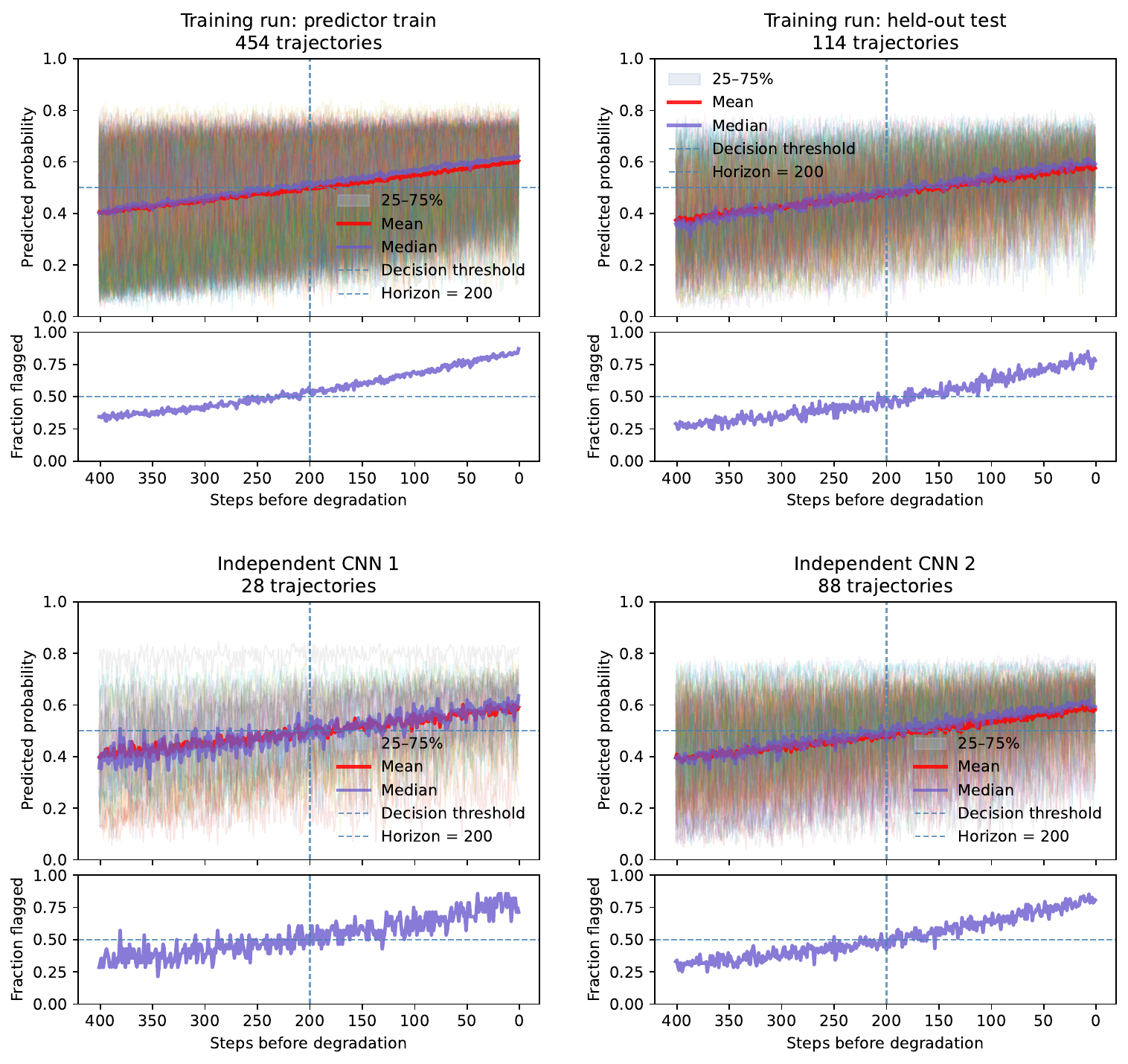}
\end{minipage}
\hfill
\begin{minipage}{0.49\textwidth}
\centering
\includegraphics[width=\linewidth]{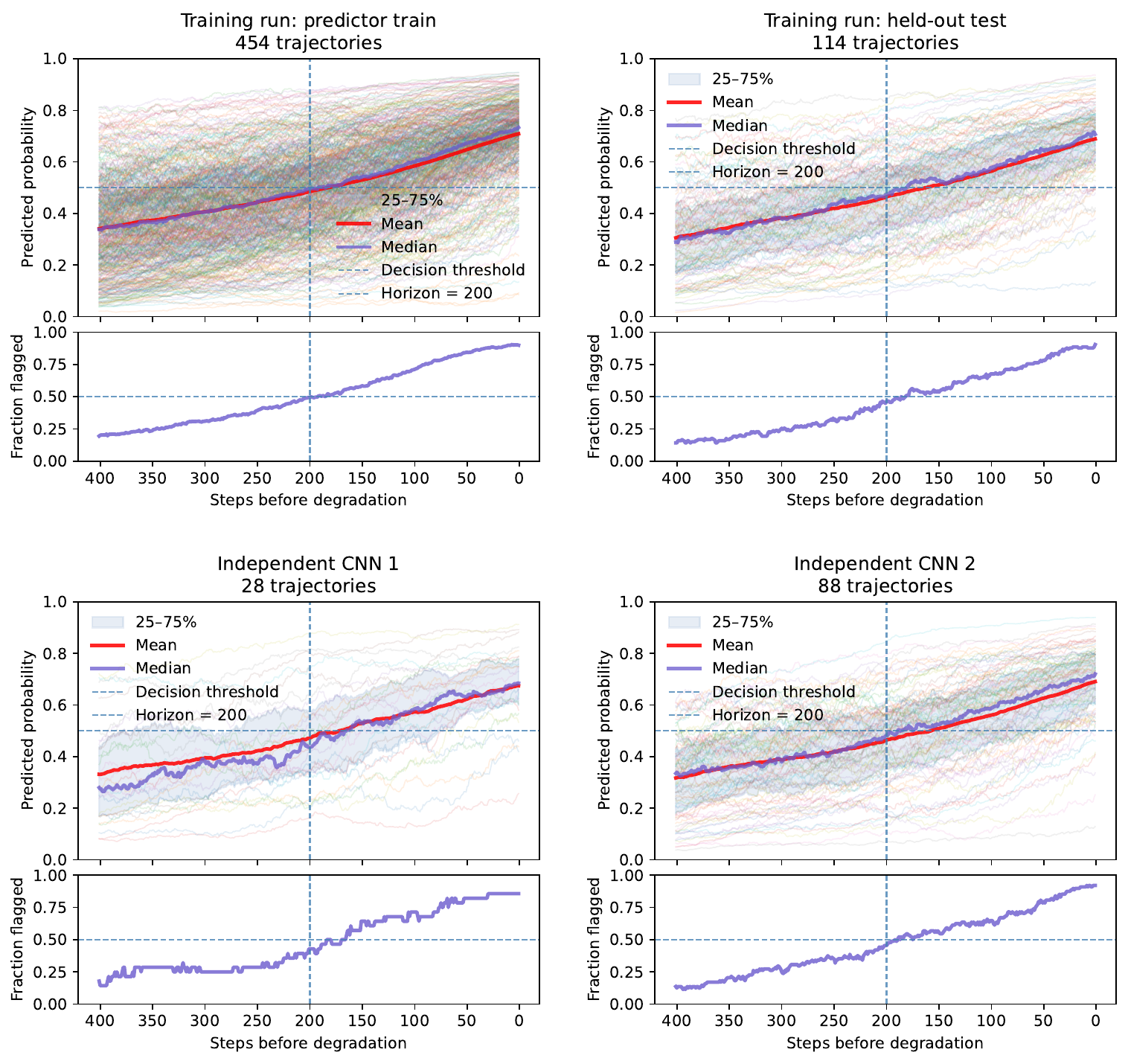}
\end{minipage}
\caption{Comparison of different temporal features on the balanced prediction datasets. Left: predictor using only the current value of each source-side metric. Right: predictor using only the mean over the preceding window.}
\label{fig:current_mean_predictors}
\end{figure}

We next compare different ways of using the recent training history while keeping all source-side metrics. Figure~\ref{fig:current_mean_predictors} compares the predictor using only the current value of each metric with the predictor using only the mean over the preceding window. As discussed in the main text, averaging over the recent history substantially improves the prediction performance.

\begin{figure}[h!]
\centering
\begin{minipage}{0.49\textwidth}
\centering
\includegraphics[width=\linewidth]{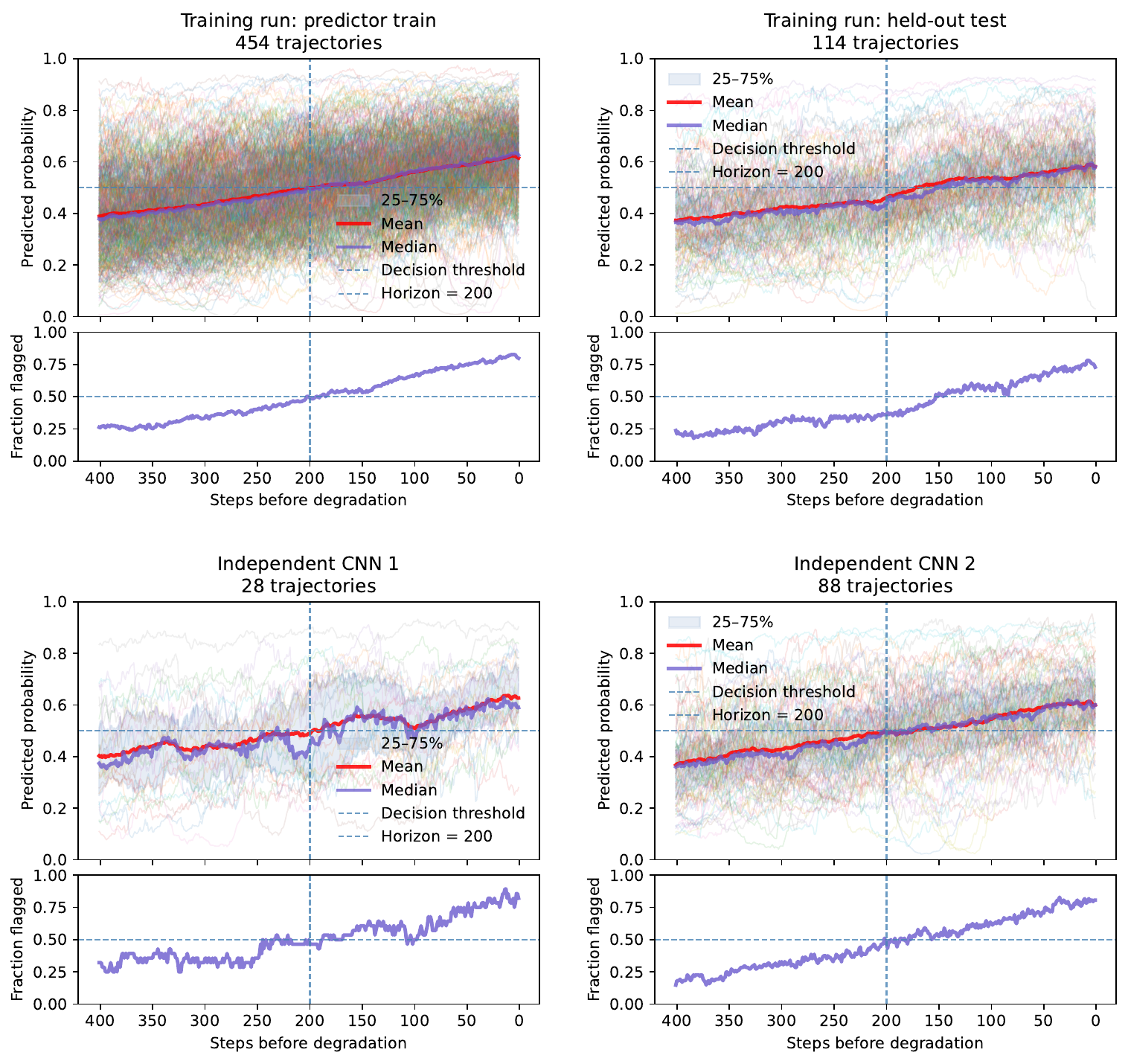}
\end{minipage}
\hfill
\begin{minipage}{0.49\textwidth}
\centering
\includegraphics[width=\linewidth]{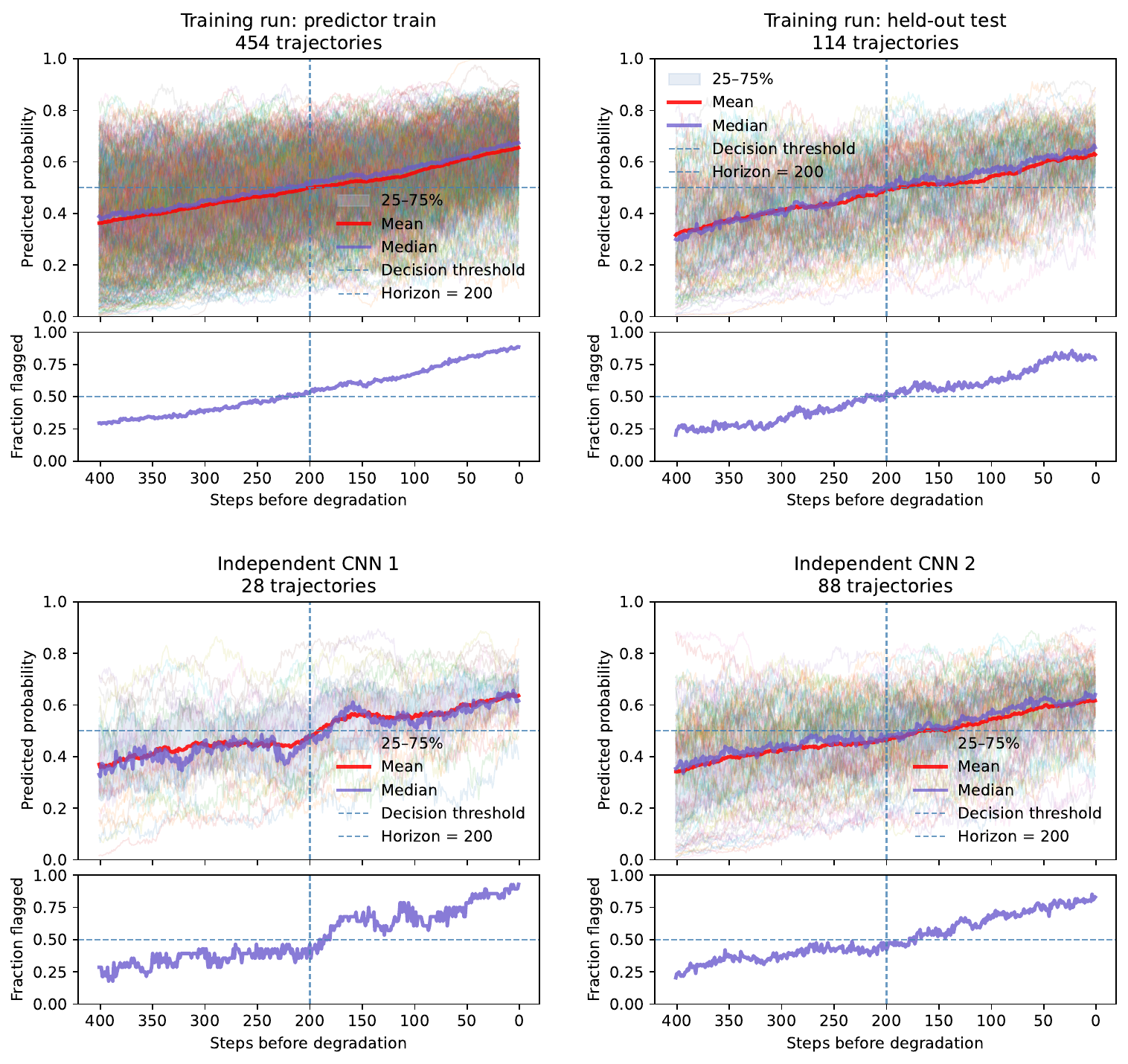}
\end{minipage}
\caption{Comparison of different groups of source-side metrics on the balanced prediction datasets, using all four temporal features. Left: parameter- and gradient-based quantities. Right: output statistics consisting of confidence, confidence on correctly classified examples, and entropy.}
\label{fig:metric_group_predictors}
\end{figure}

Finally, we separate the remaining observables into two broad groups. Figure~\ref{fig:metric_group_predictors} compares the predictor restricted to the parameter- and gradient-based quantities with the predictor restricted to the output statistics, consisting of confidence, confidence on correctly classified examples, and entropy.

\FloatBarrier
\bibliographystyle{JHEP_no_lower}
\bibliography{prospective_prediction}{}

\end{document}